\def\year{2018}\relax
\documentclass[letterpaper]{article} 
\usepackage{aaai18}  
\usepackage{times}  
\usepackage{helvet}  
\usepackage{courier}  
\usepackage{url}  
\usepackage{graphicx}  
\usepackage{epsfig}
\usepackage{amsmath}
\usepackage{amssymb}
\usepackage{color}
\usepackage{comment}

\begin{document}
%
\title{Examining CNN Representations With Respect To Dataset Bias}
\author{Quanshi Zhang$^{\dag}$, Wenguan Wang$^{\dag,\ddag}$, and Song-Chun Zhu$^{\dag}$\\
$^{\dag}$University of California, Los Angeles\qquad$^{\ddag}$Beijing Institute of Technology}

\maketitle

\begin{abstract}
Given a pre-trained CNN without any testing samples, this paper proposes a simple yet effective method to diagnose feature representations of the CNN. We aim to discover representation flaws caused by potential dataset bias. More specifically, when the CNN is trained to estimate image attributes, we mine latent relationships between representations of different attributes inside the CNN. Then, we compare the mined attribute relationships with ground-truth attribute relationships to discover the CNN's blind spots and failure modes due to dataset bias. In fact, representation flaws caused by dataset bias cannot be examined by conventional evaluation strategies based on testing images, because testing images may also have a similar bias. Experiments have demonstrated the effectiveness of our method.
\end{abstract}

\section{Introduction}

Given a convolutional neural network (CNN) that is pre-trained to estimate image attributes (or labels), how to diagnose black-box knowledge representations inside the CNN and discover potential representation flaws is a crucial issue for deep learning. In fact, there is no theoretical solution to identifying good and problematic representations in the CNN. Instead, people usually just evaluate a CNN based on the accuracy obtained using testing samples.

\begin{figure}[t]
\centering
\includegraphics[width=\linewidth]{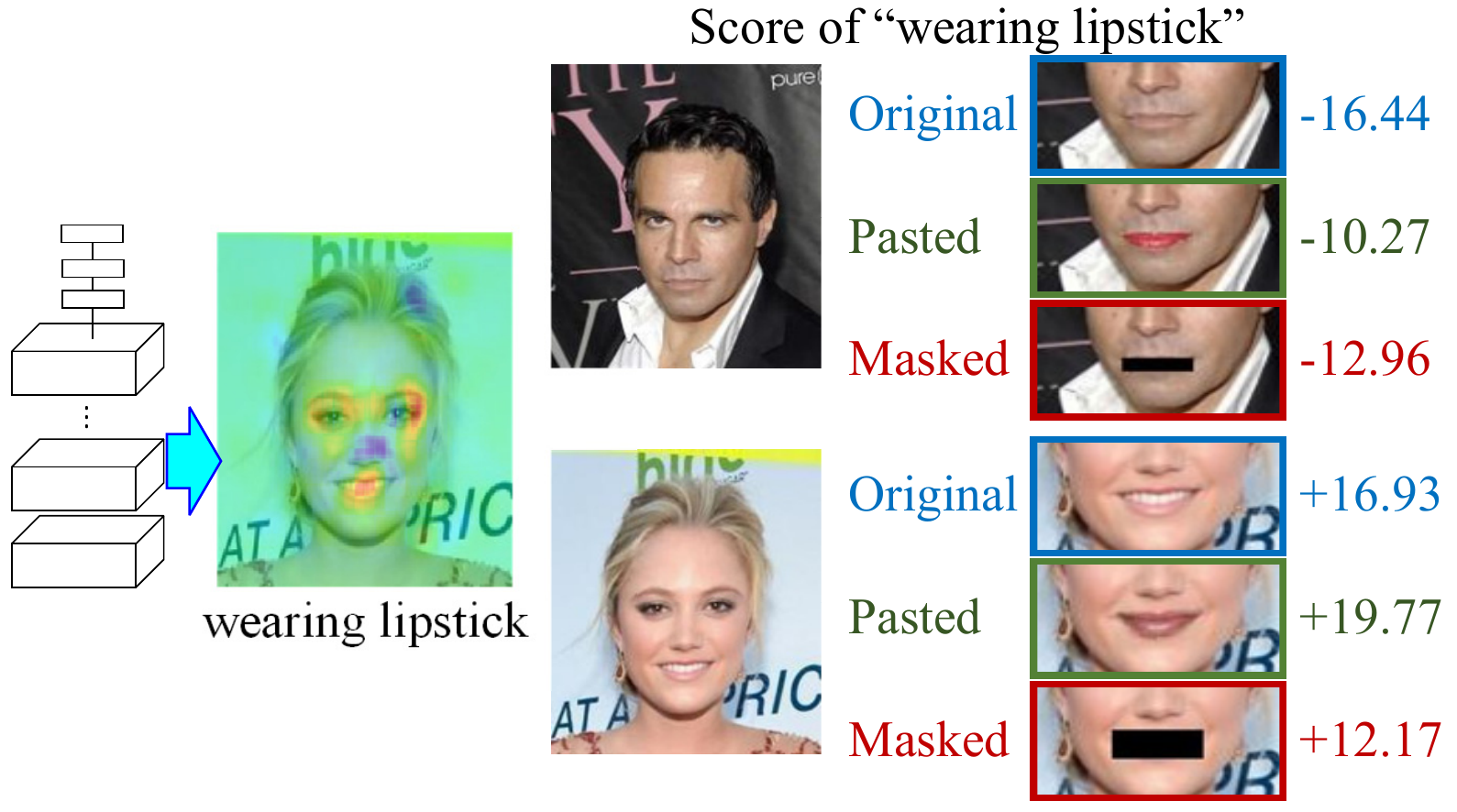}
\caption{Biased representations in a CNN. Considering potential dataset bias, a high accuracy on testing images cannot always ensure that a CNN learns correct representations. The CNN may use unreliable co-appearing contexts to make predictions. For example, we manually modify mouth appearances of two faces by masking mouth regions or pasting another mouth, but such modifications do not significantly change prediction scores for the \textit{lipstick} attribute. We show heat maps of inference patterns of the \textit{lipstick} attribute, where patterns with red/blue colors are positive/negitive with the attribute score. The CNN mistakenly considers unrelated patterns as contexts to infer the lipstick. We propose a method to automatically discover such biased representations from a CNN without any testing images.}
\label{fig:intro}
\end{figure}

In this study, we focus on representation flaws caused by potential bias in the collection of training samples~\cite{datasetBias}. As shown in Fig.~\ref{fig:intro}, if an attribute usually co-appears with certain visual features in training samples, then the CNN may be learned to use the co-appearing features to represent this attribute. When the used co-appearing features are not semantically related to the target attribute, we consider these features as biased representations. This idea is related to the disentanglement of the local, bottom-up, and top-down information components for prediction~\cite{wu2007compositional,yang2009evaluating,wu2011numerical}. We need to clarify correct and problematic contexts for prediction. CNN representations may be biased even when the CNN achieves a high accuracy on testing samples, because testing samples may have a similar bias.

In this paper, we propose a simple yet effective method that automatically diagnoses representations of a pre-trained CNN without given any testing samples. \emph{I.e.,} we only use training samples to determine the attributes whose representations are not well learned. We discover blind spots and failure modes of the representations, which can guide the collection of new training samples.

\begin{figure*}[t]
\centering
\includegraphics[width=0.9\linewidth]{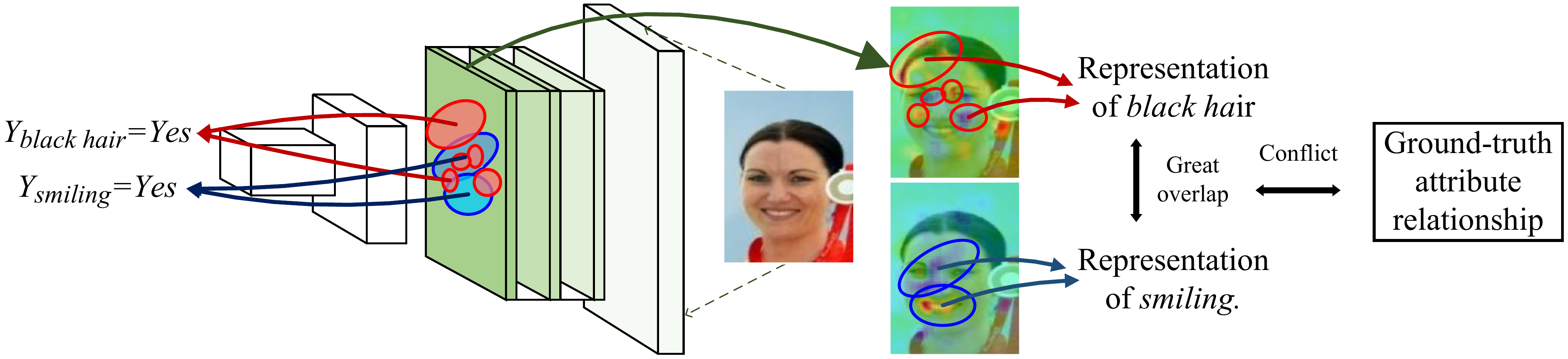}
\caption{Overview of the method. Given a biased dataset for training where the \textit{smiling} and \textit{black hair} attributes usually appear on faces with certain appearances of eyes or noses, the CNN may mistakenly use eye or nose features to represent the two attributes. Biased representations are difficult to discover when testing samples are also biased. In this study, we mine relationships between attributes. Conflicts between the mined and ground-truth relationships indicate potential representation problems.}
\label{fig:top}
\end{figure*}

\textbf{Intuition, self-compatibility of network representations: }{\verb| |} Given a pre-trained CNN and an image $I$, we use the CNN to estimate attribute $A$ for $I$. We also mine inference patterns\footnote[1]{We regard a neural pattern as a group of units in a channel of a conv-layer's feature map, which are activated and play a crucial role in the estimation of the attribute $A$.} of the estimation result, which are hidden in conv-layers of the CNN. We can regard the mined inference patterns as exact representations of the attribute $A$ in the CNN. Then, based on inference patterns, we compute the relationship between each pair of attributes {$(A_{i},A_{j})$}, \emph{i.e.} identifying whether {$A_{i}$} is positively/negatively/not related to {$A_{j}$}.

The intuition is simple, \emph{i.e.} according to human's common sense, we set up several ground-truth relationships between some pairs of attributes as rules to diagnose CNN representations. The mined attribute relationships should well fit the ground truth; otherwise, the representation is probably not well learned. Let us take a CNN that is learned to estimate face attributes for example. As shown in Fig.~\ref{fig:top}, the \textit{smiling} attribute is supposed to be represented by features (patterns), which appear on the mouth region in conv-layers. Whereas, the \textit{black hair} attribute should be inferred by features extracted from hairs. Therefore, the attribute relationship ``\textit{smiling} is not related to the \textit{black hair}'' is trustworthy enough to become a ground truth. However, the CNN may use eye/nose features to represent the two attribute, because these attributes always co-appear with specific eye/nose appearances in a biased dataset. Thus, we will mine a specific relationship between the two attributes, which conflicts with the ground truth.

\textbf{Our method: }{\verb| |} Given a pre-trained CNN, we mine relationships between each pair of attributes according to their inference patterns. Then, we annotate some ground-truth attribute relationships. For example, the \textit{heavy makeup} attribute is positively related to the \textit{attractive} attribute; \textit{black hair} and \textit{smiling} are not related to each other. We compute the Kullback-Leibler (KL) divergence between the mined relationships and ground-truth relationships to discover attributes that are not well learned, including both blind spots and failure modes of attribute representations.

In fact, how to define ground-truth relationships is still an open problem. We can ask different people to label attribute relationships in their personal opinions to approach the ground truth. More importantly, our method is compatible with various types of ground-truth distributions. People can define their ground truth \emph{w.r.t.} their tasks as constraints to examine the network. Thus, our method is a flexible and convincing way to discover representation bias at the level of human cognition.

The annotation cost of our method is usually much lower than end-to-end learning of CNNs. Our annotation cost is $O(n^2)$, where $n$ denotes the number of attribute outputs. In contrast, it usually requires thousands or millions of samples to learn a new CNN in real applications.

\textbf{Why is the proposed method important? }{\verb| |} As a complement to using testing samples for evaluation, our zero-shot diagnosis of a CNN is of significant values in applications:
\begin{itemize}
\item A high accuracy on potentially biased testing samples cannot prove correct representations of a CNN.
\item Potential bias cannot be fully avoided in most datasets. Especially, some attributes (\emph{e.g.} \textit{smiling}) mainly describe specific parts of images, but the dataset~\cite{CelebA,SUNAttr} only provides image-level annotations of attributes for supervision without specifying regions of interests, which makes the CNN more sensitive to dataset bias.

    More crucially, the level of representation bias is not necessary to be proportional to the dataset bias level. We need to diagnose the actual CNN representations.
\item In conventional studies, correcting representation flaws caused by either dataset bias or the over-fitting problem is a typical long-tail problem. If we blindly collect new training samples without being aware of failure modes of the representation, it would require massive new samples to overcome the bias problem. Our method provides a new perspective to solve the long-tail problem.
\item Unlike methods of CNN visualization/analysis~\cite{CNNVisualization_1,CNNVisualization_2,CNNVisualization_3,trust} that require people to one-by-one check the representation of each image, our method discovers all biased representations in a batch.
\end{itemize}

\textbf{Contribution:}{\verb| |} In this study, to the best of our knowledge, we, for the first time, propose a method to discover potentially biased representations hidden in a pre-trained CNN without testing samples. Our method mines blind spots and failure modes of a CNN in a batch manner, which can guide the collection of new samples. Experiments have proved the effectiveness of the proposed method.

\section{Related work}

\textbf{Visualization of CNNs: }{\verb| |} In order to open the black box of a CNN, many methods~\cite{CNNVisualization_1,CNNVisualization_2,CNNVisualization_3,CNNVisualization_5,CNNFeatureMining,FeaVisual} have been developed to visualize and analyze patterns of response units in a CNN. Some methods~\cite{CNNVisualization_1,CNNVisualization_2,CNNVisualization_3} back-propagate gradients \emph{w.r.t.} a given unit to pixel values of an image, in order to obtain an image that maximizes the score of the unit. These techniques mainly visualize simple patterns. As mentioned in \cite{attribute}, attributes are an important perspective to model images, but it is difficult to visualize a complex attribute (\emph{e.g.} the \textit{attractive} attribute).

Given a feature map produced by a CNN, Dosovitskiy \emph{et al.}~\cite{FeaVisual} trained a new up-convolutional network to invert the feature map to the original image. Similarly, this approach was not designed for the visualization of a single attribute output.

\textbf{Interpreting semantic meanings of CNNs: }{\verb| |} Going beyond the ``passive'' visualization of neural patterns, some studies ``actively'' retrieve mid-level patterns from conv-layers, which potentially corresponds to a certain object/image part. Zhou~\emph{et al.}~\cite{CNNSemanticDeep,CNNSemanticDeep2} mined patterns for ``scene'' semantics from feature maps of a CNN. Simon~\emph{et al.} discovered objects~\cite{ObjectDiscoveryCNN_2} from CNN feature maps in an unsupervised manner, and retrieved patterns for object parts in a supervised fashion~\cite{CNNSemanticPart}. Zhang~\emph{et al.}~\cite{CNNAoG} used a graphical model to organize implicit mid-level patterns mined from a CNN, in order to explain the pattern hierarchy inside conv-layers in a weakly-supervised manner. \cite{interpretVQA_grad} used a gradient-based method to interpret visual question-answering models. Zhang~\emph{et al.}~\cite{explanatoryGraph} transformed CNN representations to an explanatory graph, which represents the semantic hierarchy hidden inside a pre-trained CNN.

\textbf{Model diagnosis: }{\verb| |} Many methods have been developed to diagnose representations of a black-box model. \cite{blackBoxKeyFeature} extracted key features for model outputs. The LIME method proposed by Ribeiro \emph{et al.}~\cite{trust} and gradient-based visualization methods~\cite{visualCNN_grad,visualCNN_grad_2} extracted image regions that were responsible for each network output, in order to interpret the network representation.

Unlike above studies diagnosing representations for each image one by one, many approaches aim to evaluate all potential attribute/label representations for all images in a batch. Lakkaraju \emph{et al.}~\cite{banditUnknown} and Zhang \emph{et al.}~\cite{DeepQA,interactiveAOG_arXiv} explored unknown knowledge hidden in CNNs via active annotations and active question-answering. Methods of \cite{failureMode,failureMode2} computed the distributions of a CNN's prediction errors among testing samples, in order to summarize failure modes of the CNN. However, we believe that compared to \cite{failureMode,failureMode2}, it is of larger value to explore evidence of failure cases from mid-layer representations of a CNN. \cite{rightReason} required people to label dimensions of input features that were related to each output according to common sense, in order to learn a better model. Hu \emph{et al.}~\cite{LogicRuleNetwork} designed some logic rules for network outputs, and used these rules to regularize the learning of neural networks. In our research, we are inspired by Deng \emph{et al.}~\cite{labelGraph}, which used label graph for object classification. We use ground-truth attribute relationships as logic rules to harness mid-layer representations of attributes. \cite{wu2007compositional,yang2009evaluating,wu2011numerical} tried to isolate and diagnose information from local, bottom-up, or top-down inference processes. More specially, \cite{wu2011numerical} proposed to separate implicit local representations and explicit contextual information used for prediction. Following this direction, this is the first study to diagnose unreliable contextual information from CNN representations \emph{w.r.t.} dataset bias.

\textbf{Active learning: }{\verb| |} Active learning is a well-known strategy for detecting ``unknown unknowns'' of a pre-trained model. Given a large number of unlabeled samples, existing methods mainly select samples on the decision boundary~\cite{activeDPMGrauman} or samples that cannot well fit the model~\cite{Active2,DeepQA}, and require human users to label these samples.

Compared to active-learning approaches, our method does not require any additional unlabeled samples to test the model. More crucially, our method looks deep inside the representation of each attribute to mine attribute relationships; whereas active learning is closer to black-box testing of model performance. As discussed in \cite{MachineTeachingActive}, unless the initial training set contains at least one sample in each possible mode of sample features, active learning may not exhibit high efficiency in model refinement.

\section{Algorithm}
\subsection{Problem description}
\label{sec:description}

We are given a CNN that is trained using a set of images {${\bf I}$} with attribute annotations. The CNN is designed to estimate $n$ attributes of an image, denoted by {$A_1,A_2,\ldots,A_{n}$}. Meanwhile, we also have a certain number of ground-truth relationships between different attributes, denoted by a relationship graph {$G^{*}=(\{A_{i}\},{\bf E}^{*})$}. Each edge {$(A_{i},A_{j})\in{\bf E}^{*}$} represents the relationship between {$A_{i}$} and {$A_{j}$}. Note that it is \textbf{not} necessary for {$G^{*}$} to be a complete graph. We only select trustworthy relationships as ground truth. The goal is to identify attributes that are not well learned and to discover blind spots and failure modes in attribute representation.

Given an image {$I\in{\bf I}$}, let {$Y_{i}^{I}$} and {$Y^{I,*}_{i}$} denote the attribute value of {$A_{i}$} estimated by the CNN and the ground-truth annotation for $A_{i}$. In order to simplify the notation, we omit superscript $I$ and use notations of {$Y_{i}$} and {$Y^{*}_{i}$} in most sections, except in Section~\ref{sec:inference}.

In different applications, people use multiple ways to define attributes (or labels), including binary attributes ({$Y_{i}\in\{-1,+1\}$}) and continuous attributes (\emph{e.g.} {$Y_{i}\in[-1,+1]$} and {$Y_{i}\in(-\infty,+\infty)$}). We can normalize all these attributes to the range of {$Y_{i}\in(-\infty,+\infty)$} for simplification\footnote[2]{Given annotations of continuous attributes $Y^{*}_{i}\in(-\infty,+\infty)$, we can define L-2 norm loss $L(Y_{i},Y^{*}_{i})=(Y^{*}_{i}-Y_{i})^2$ to train the CNN. Given annotations of binary attributes for training $Y^{*}_{i}\in\{-1,+1\}$, we can use the logistic log loss $L(Y_{i},Y^{*}_{i})=\log(1+\exp(-Y_{i}\cdot Y^{*}_{i}))$ to train the CNN. In this way, $Y_{i}$ can be considered as an attribute estimation whose range is $(-\infty,+\infty)$.}. To simplify the introduction, without loss of generality, we consider {$Y^{*}_{i}>0$} as the existence of a certain attribute {$A_{i}$}; otherwise not. Consequently, we flip the signs of some ground-truth annotations to ensure that we use positive values, rather than negative values, to represent the activation of {$A_{i}$}.


\subsection{Mining attribute relationships}
\label{sec:inference}

\begin{figure*}[t]
\centering
\includegraphics[width=0.9\linewidth]{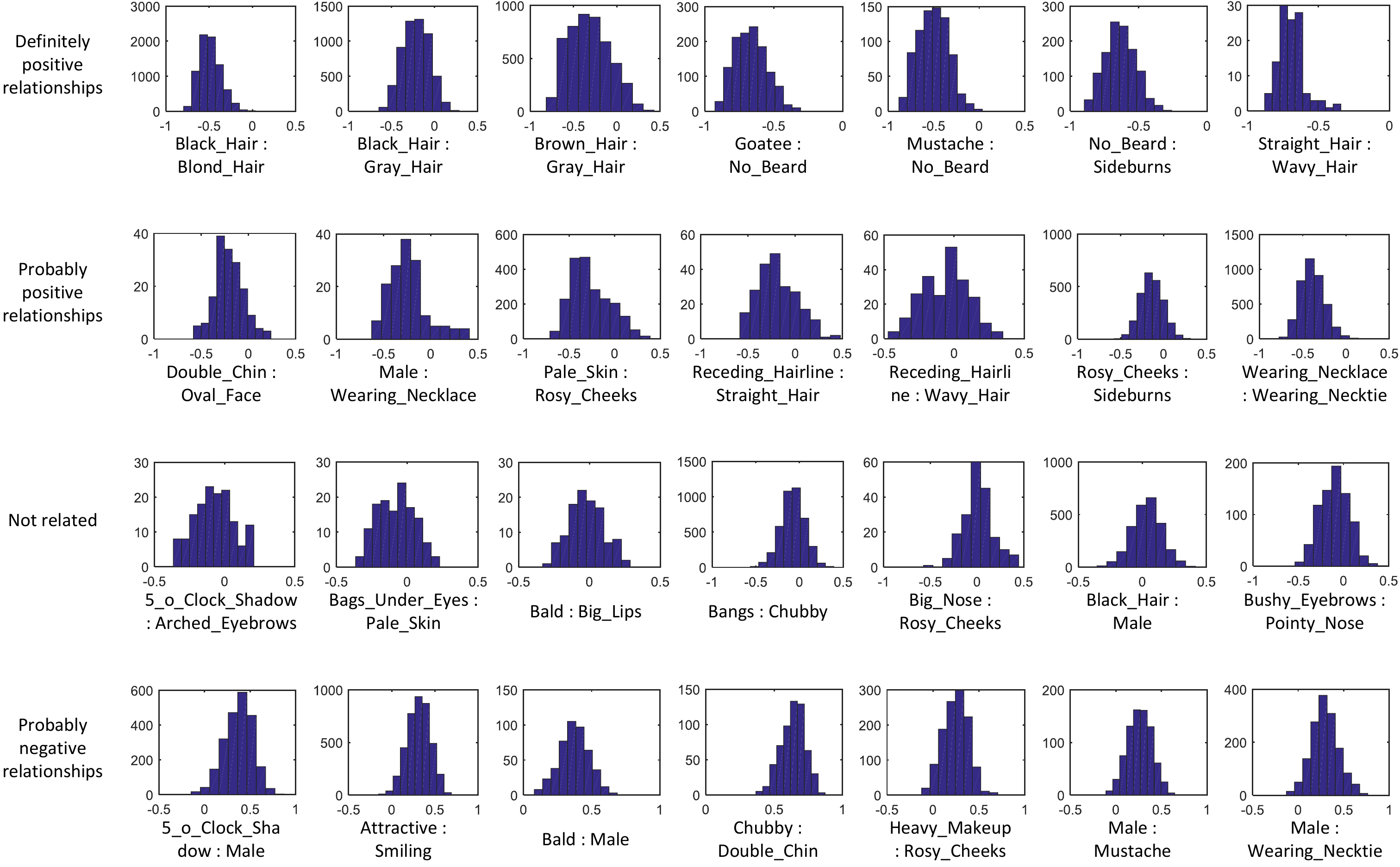}
\caption{Histograms of $\varpi_{ij}$ to describe distributions of ${\bf Q}(\varpi_{ij}|A_{i},A_{j})$ for different pairs of attributes. The horizontal axis indicates the value of $\varpi_{ij}$. Note that the vertical axis indicates the number of samples in the histogram, which is not the density of ${\bf Q}(\varpi_{ij}|A_{i},A_{j})$. In the first, second, third, and fourth rows, we show the mined distributions for attribute pairs that are labeled with ``definitely positive relationships,'' ``probably positive relationships,'' ``not-related relationships,'' and ``probably negative relationships,'' respectively.}
\label{fig:hist}
\end{figure*}

\textbf{Attribute representation:}{\verb| |} Given an image {$I\in{\bf I}$} and a target attribute {$A_{i}$}, we select the feature map {${\bf x}^{I}$} of a certain conv-layer of the CNN to represent {$A_{i}$} and compute {$Y_{i}^{I}$}. Since the CNN conducts a series of convolution and ReLu operations on {${\bf x}^{I}$} to compute {$Y_{i}^{I}$}, we can approximate {$Y_{i}^{I}$} as a linear combination of neural activations in {${\bf x}^{I}$}.
\begin{equation}
Y_{i}^{I}\approx({\bf v}_{i}^{I})^{T}{\bf x}^{I}+\beta_{i}^{I},\qquad{\bf v}_{i}^{I}={\boldsymbol\rho}_{i}\circ{\boldsymbol\nu}_{i}^{I}
\label{eqn:linear}
\end{equation}
where {${\bf v}_{i}^{I}$} denotes a weight vector, and {$\beta_{i}^{I}$} is a scalar for bias.

In the above equation, parameters {${\boldsymbol\nu}_{i}^{I}$} and {$\beta_{i}^{I}$} reflect inherent piecewise linear representations of {$Y_{i}^{I}$} inside the CNN, whose values have been fixed when the CNN and the image are given. We will introduce the estimation of {${\boldsymbol\nu}_{i}^{I}$} and {$\beta_{i}^{I}$} later. The target parameter here is {${\boldsymbol\rho}_{i}\in\{0,1\}^{N}$}, which is a sparse mask vector. It means that we select a relatively small number of reliable neural activations from {${\bf x}^{I}$} as inference patterns of $A_{i}$ and filters out noises. $\circ$ denotes element-wise multiplication between vectors. We can regard {${\boldsymbol\rho}_{i}$} as a prior spatial distribution of neural activations that are related to attribute $A_{i}$. For example, if $A_{i}$ represents an attribute for noses, then we expect {${\boldsymbol\rho}_{i}$} to mainly represent nose regions. Note that except {${\boldsymbol\rho}_{i}$}, parameters {${\boldsymbol\nu}_{i}^{I}$} and {$\beta_{i}^{I}$} are only oriented to image $I$ due to ReLu operations in the CNN.

We can compute the inherent piecewise linear gradient \emph{w.r.t.} {${\bf x}^{I}$}, \emph{i.e.} {${\boldsymbol\nu}_{i}^{I}$} via gradient back propagation.
\begin{equation}
{\boldsymbol\nu}_{i}^{I}=\frac{\partial Y_{i}}{\partial{\bf x}}{\Big|}_{{\bf x}={\bf x}^{I}}=\frac{\partial Y_{i}}{\partial{\bf x}_{M}}\frac{\partial{\bf x}_{M}}{\partial{\bf x}_{M-1}}\cdots\frac{\partial{\bf x}_{m+2}}{\partial{\bf x}_{m+1}}\frac{\partial{\bf x}_{m+1}}{\partial{\bf x}}{\Big|}_{{\bf x}={\bf x}^{I}}
\label{eqn:grad}
\end{equation}
where the CNN contains $M$ conv-layres (including fully-connected layers), and {${\bf x}_{k}$} denotes the output of the $k$-th conv-layer (${\bf x}\overset{\text{def}}{=}{\bf x}_{m}$ corresponds to the $m$-th conv-layer). We can further compute the value of {$\beta_{i}^{I}$} based on the full representation without pattern selection {$Y_{i}^{I}=({\boldsymbol\nu}_{i}^{I})^{T}{\bf x}^{I}+\beta_{i}^{I}$}.

Inspired by the LIME method~\cite{trust}, the loss of mining inference patterns is similar to a Lasso selection:
\begin{equation}
\hat{\boldsymbol\rho}_{i}=\underset{{\boldsymbol\rho}_{i}}{\arg\!\min}\;{\bf E}_{I\in{\bf I}}\big[{\mathcal L}(Y_{i}^{I},{\boldsymbol\rho}_{i})\big]+{\bf L}({\boldsymbol\rho}_{i})
\label{eqn:loss}
\end{equation}
where {${\mathcal L}(Y_{i}^{I},{\boldsymbol\rho}_{i})$} measures the fidelity of the representation on image $I$, and {${\bf L}({\boldsymbol\rho}_{i})$} denotes the representation complexity. We can simply formulate {${\mathcal L}(Y_{i}^{I},{\boldsymbol\rho}_{i})=[({\bf v}_{i}^{I})^{T}{\bf x}^{I}+\beta_{i}^{I}-Y_{i}^{I}]^2$}, and {${\bf L}({\boldsymbol\rho}_{i})=\lambda\Vert{\boldsymbol\rho}_{i}\Vert_1$}, where {$\Vert\cdot\Vert_1$} denotes L-1 norm, and {$\lambda$} is a constant. Based on the above equation, {$\hat{\boldsymbol\rho}_{i}$} can be directly estimated using a greedy strategy.

\textbf{Attribute relationships:}{\verb| |} For each pair of attributes {$A_{i}$} and {$A_{j}$}, we define a cosine distance {$\varpi_{ij}^{I}\overset{\text{def}}{=}\frac{({\bf v}^{I}_{j})^{T}{\bf v}^{I}_{i}}{\Vert{\bf v}^{I}_{j}\Vert\Vert{\bf v}^{I}_{i}\Vert}$} to represent their attribute relationship. If {$A_{i}$} and {$A_{j}$} are positively related, {${\bf v}_{i}$} will approximate to {${\bf v}_{j}$}, \emph{i.e.} {$\varpi_{ij}^{I}$} will be close to 1. Similarly, if {$A_{i}$} and {$A_{j}$} are negatively related, then {$\varpi_{ij}^{I}$} will be close to -1. If {$A_{i}$} and {$A_{j}$} are not closely related, then {${\bf v}_{j}$} and {${\bf v}_{i}$} will be almost orthogonal, thus {$\varpi_{ji}^{I}\approx0$}.


The actual representation of an attribute in a CNN is highly non-linear, and the linear representation in Eq.~(\ref{eqn:linear}) is just a local mode oriented to a specific image $I$. When we compute the gradient {${\boldsymbol\nu}_{i}^{I}=\frac{\partial Y_{i}}{\partial{\bf x}^{I}}$}, the ReLu operation blocks irrelevant information in gradient back-propagation, thereby obtaining a local linear representation. It is possible to cluster {${\boldsymbol\nu}_{i}^{I}$} of different images into several local modes of the representation. Expect extreme cases mentioned in \cite{neuralGap}, these local modes are robust to most small perturbations in the image $I$.

\subsection{Diagnosis of CNN representations}

Given each image {$I\in{\bf I}$}, we compute {$\varpi_{ij}^{I}$} to represent the relationship between {$A_{i}$} and {$A_{j}$} \emph{w.r.t.} the image $I$. In this way, we use the distribution of {$\varpi_{ij}^{I}$} among all training images in {${\bf I}$}, denoted by {${\bf Q}(\varpi_{ij}|A_{i},A_{j})$}, to represent the overall attribute relationship\footnote[3]{Without loss of generality, we modify attribute annotations to ensure {$Y_{i}^{*}=+1$} rather than {$Y_{i}^{*}=-1$} to indicate the existence of a certain attribute. We find that the CNN mainly extracts common patterns from positive samples as inference patterns to represent each attribute. Thus, we compute distributions {${\bf P}$} and {${\bf Q}$} for {$(A_{i},A_{j})$} among the samples in which either {$Y_{i}^{*}=+1$} or {$Y_{j}^{*}=+1$}. Similarly, in Experiment 3, we also ignored samples with {$Y_{i}^{*}=Y_{j}^{*}=-1$} to compute the entropy for the competing method.}. Fig.~\ref{fig:hist} shows the mined distributions of {${\bf Q}(\varpi_{ij}|A_{i},A_{j})$} for different pairs of attributes.


Besides the observation distribution {${\bf Q}$}, we also manually annotate a number of ground-truth attribute relationships {$G^{*}$}, and define a distribution for each ground-truth attribute relationship {${\bf P}(\varpi_{ij}|A_{i},A_{j})$}. People can label several types of ground-truth relationships for {$(A_{i},A_{j})\in{\bf E}^{*}$}, {$l_{ij}\in{\bf L}=\{L_1,L_2,\ldots\}$}, to supervise the diagnosis of CNN representations. Let {$(A_{i},A_{j})\in{\bf E}^{*}$} be labeled with {$l_{ij}=L^{*}$}. We assume the ground-truth distribution {${\bf P}(\varpi_{ij}|A_{i},A_{j})\sim{\mathcal N}(\mu_{L^{*}},\sigma_{L^{*}}^2)$} follows a Gaussian distribution. We assume most pairs of attributes are well learned, so we can compute {$\mu_{L^{*}}$} and {$\sigma_{L^{*}}^2$} as the mean and the variation of {$\varpi_{ij}$}, respectively, among all pairs of attributes that are labeled with {$L^{*}$}. In this way, biased representations correspond to outliers of {$\varpi_{ij}$} \emph{w.r.t} the ground-truth distribution.

\begin{figure*}[t]
\centering
\includegraphics[width=0.9\linewidth]{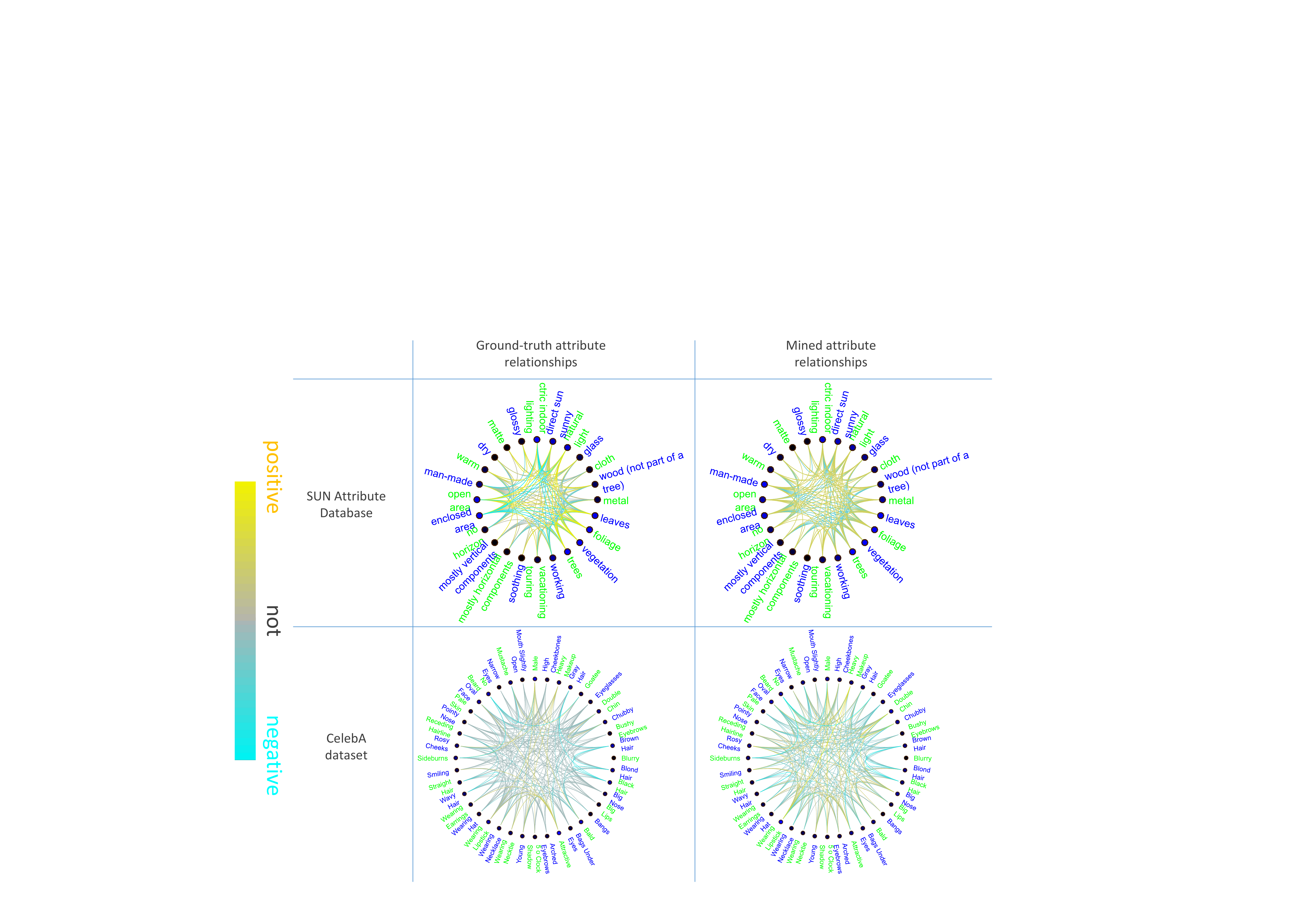}
\caption{The mined and ground-truth attribute relationships. We mine attribute relationships based on CNNs that are trained using the CelebA dataset and the SUN Attribute database. Most attributes are well learned, so most of the mined relationships well fit the ground-truth. The edge color indicates {${\bf E}_{{\bf P}\,\textrm{or}\,{\bf Q}}[\varpi_{ij}]$}. The yellow/gray/cyan color indicates the positive/no/negative relationship between attributes. For clarity, we randomly draw 300 edges of the relationship graph.}
\label{fig:graph}
\end{figure*}

\begin{figure*}[t]
\centering
\includegraphics[width=0.9\linewidth]{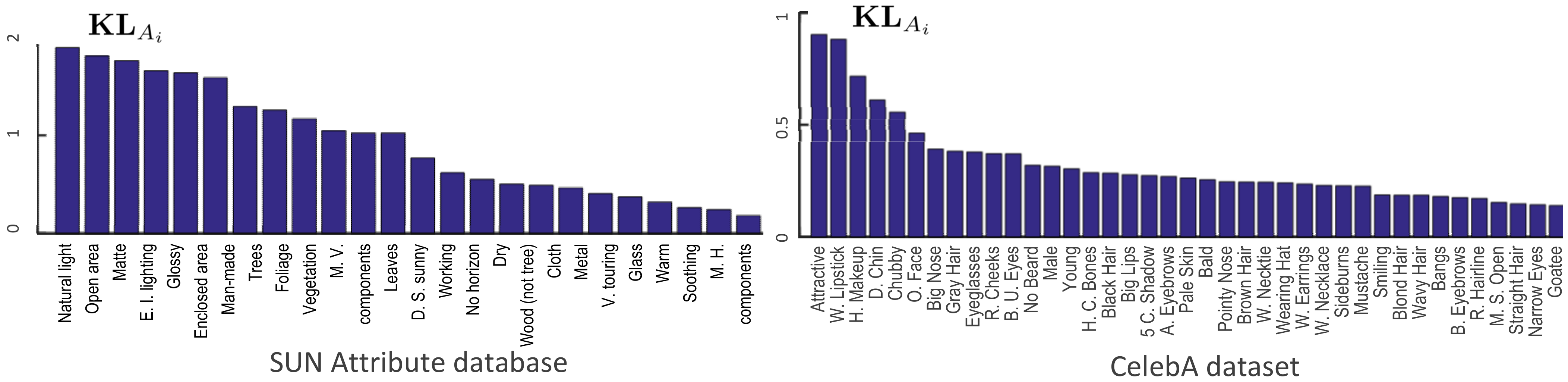}
\caption{The mined KL-divergences {${\bf KL}_{A_{i}}$} of different attributes. Attributes with lower KL-divergences are better learned.}
\label{fig:graph_kl}
\end{figure*}

We can compute the KL-divergence between {${\bf P}$} and {${\bf Q}$}, {${\bf KL}({\bf P}\Vert{\bf Q})$}, to discover biased representations.
\begin{small}\begin{eqnarray}
{\bf KL}_{A_{i}A_{j}}=\int_{\Omega}{\bf P}(\varpi_{ij}|A_{j},A_{i})\log\frac{{\bf P}(\varpi_{ij}|A_{j},A_{i})}{{\bf Q}(\varpi_{ij}|A_{j},A_{i})}{\bf d}\varpi_{ij}\\
\!{\bf KL}_{A_{i}}\!\!=\!\!\!\!\!\!\!\!\!\!\!\!\!\sum_{j:(A_{i},A_{j})\in{\bf E}^{*}}\!\!\!\int_{\Omega}{\bf P}(\varpi_{ij},A_{j}|A_{i})\log\frac{{\bf P}(\varpi_{ij},A_{j}|A_{i})}{{\bf Q}(\varpi_{ij},A_{j}|A_{i})}{\bf d}\varpi_{ij}\!\!\!\!\!\\
=\!\!\!\sum_{j:(A_{i},A_{j})\in{\bf E}^{*}}\!\!P(A_{j}|A_{i}){\bf KL}_{A_{i}A_{j}}\nonumber
\end{eqnarray}\end{small}
where {$P(A_{j}|A_{i})=1/\textrm{deg}(A_{i})$} is a constant given the degree of {$A_{i}$}. We approximately set {$\Omega\!=\![-1,1]$}, because {${\bf P}(\varpi_{ij}|A_{j},A_{i})\approx0$} when {$\vert\varpi_{ij}\vert>1$} in real applications. We believe that if {${\bf KL}_{A_{i}}$} is high, {$A_{i}$} is probably not well learned.

\textbf{Blind spots \& failure modes: }{\verb| |} Each pair of attributes {$(A_{i},A_{j})\in{\bf E}^{*}$} with a high {${\bf KL}_{A_{i}A_{j}}$} may have two alternative explanations. The first explanation is that {$(A_{i},A_{j})$} represents a blind spot of the CNN. \emph{I.e.} {$(A_{i},A_{j})$} should be positively/negatively related to each other according to the ground-truth, but the CNN has not learned many inference patterns that are shared by both {$A_{i}$} and {$A_{j}$}. In this case, the CNN does not encode the inference relationship between {$A_{i}$} and {$A_{j}$}.

The alternative explanation is that {$(A_{i},A_{j})$} represents a failure mode. If the mined relationship is that {$A_{i}$} is strongly positively related to {$A_{j}$}, which conflicts with the ground-truth relationship. Then, samples with opposite ground-truth annotations for {$A_{i}$} and {$A_{j}$}, {$Y_{i}^{*}\cdot Y_{j}^{*}\!<\!0$} may correspond to a failure mode in attribute estimation. Note that these samples belong to two modes, \emph{i.e.} the modes of {$Y_{i}^{*}>0,Y_{j}^{*}\!<\!0$} and {$Y_{i}^{*}<0,Y_{j}^{*}\!>\!0$}. We simply select the mode with fewer samples as a failure mode. Similarly, if the CNN incorrectly encodes a negative relationship between {$(A_{i},A_{j})$}, then we select a failure mode from candidates of {$(Y_{i}^{*}\!>\!0,Y_{j}^{*}\!>\!0)$} and {$(Y_{i}^{*}\!<\!0,Y_{j}^{*}\!<\!0)$}.

In practise, we determine blind spots and failure modes as follows. Given a pair of attributes {$(A_{i},A_{j})$} with a high {${\bf KL}_{A_{i}A_{j}}$}, if {$\vert{\bf E}_{I}[\varpi_{ij}^{I}]\vert\!<\!0.2$} and {$\vert{\bf E}_{I}[\varpi_{ij}^{I}]-\mu_{l_{ij}}\vert>0.2$}, then {$(A_{i},A_{j})$} correspond to a blind spot. If {$\vert{\bf E}_{I}[\varpi_{ij}^{I}]\vert\!>\!0.2$} and {$\vert{\bf E}_{I}[\varpi_{ij}^{I}]-\mu_{l_{ij}}\vert\!>\!0.2$}, we extract a failure mode from {$(A_{i},A_{j})$}.

\section{Experiments}

\textbf{Dataset: }{\verb| |} We tested the proposed method on the Large-scale CelebFaces Attributes (CelebA) dataset~\cite{CelebA} and the SUN Attribute database~\cite{SUNAttr}. The CelebA dataset contains more than 200K celebrity images, each with 40 attribute annotations. In order to simplify the story, we first used annotations of face bounding boxes provided in the dataset to crop face regions from original images, and then used the cropped faces as input to learn a CNN. The SUN database contains 14K scene images with 102 attributes, but most attributes only appear in very few images. Thus, we selected 24 attributes with minimum scores of {$\max(\#(Y_{i}^{*}>0),\#(Y_{i}^{*}<0))$} as target attributes for experiments, where {$\#(Y_{i}^{*}>0)$} denotes the number of positive annotations of {$A_{i}$} among all images. Furthermore, in the SUN dataset, value ranges for ground-truth attribute annotations are {$Y_{i}^{*}\in[0,1]$}. We modified the ground-truth to binary annotations {$Y_{i}^{*,new}=sign(Y_{i}^{*}-0.5)$} for simplicity.

\begin{figure*}[t]
\centering
\includegraphics[width=0.85\linewidth]{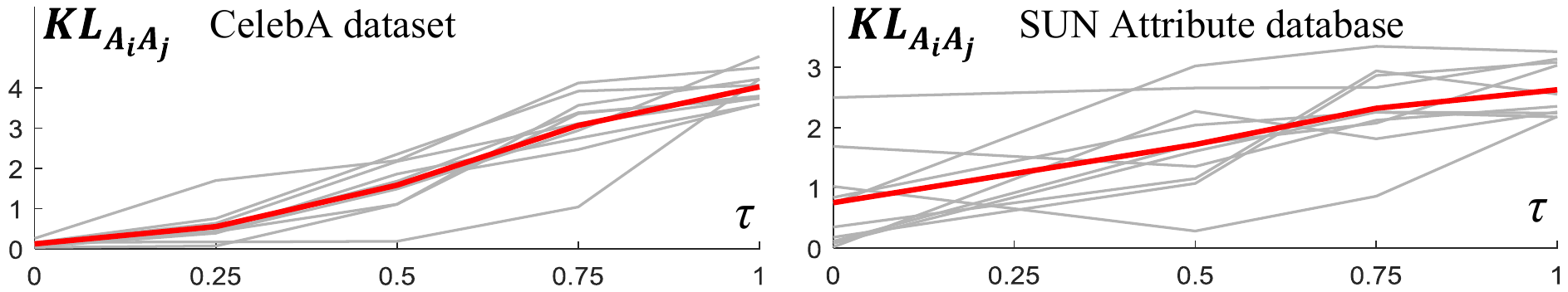}
\caption{KL divergence of attribute relationships computed on datasets that are biased at different levels $\tau$. Red curves show the average performance.}
\label{fig:kl}
\end{figure*}

\textbf{Implementation details: }{\verb| |} In this study, we used the AlexNet~\cite{CNNImageNet} as the target CNN, which contains five conv-layers and three fully-connected layers. We tracked inference patterns of an attribute through different conv-layers, and we used inference patterns in the first conv-layer for CNN diagnosis. It is because that feature maps in lower conv-layers have higher resolutions and that inference patterns in lower conv-layers are better localized than higher conv-layers. Although low-layer patterns mainly represent simple shapes (\emph{e.g.} edges), edges on \textit{black hairs} and edges describing \textit{smiling} should be localized at different positions.

For the CelebA dataset, we defined five types of attribute relationships\footnote[4]{``Definitely negative'' is referred to as exclusive attributes, \emph{e.g.} \textit{black hair} and \textit{blond hair}. Whereas, ``probably'' means a high probability. For example, a \textit{heavy makeup} person is probably \textit{attractive}.}, \emph{i.e.} {$l_{ij}\in\{\textrm{definitely negative},$ $\textrm{probably negative},$ $\textrm{not related},$ $\textrm{probably positive},$ $\textrm{definitely positive}\}$}{\color{red}{\footnotemark[4]}}. We obtained {$\mu_{\textrm{definitely positive}}>\mu_{\textrm{probably positive}}>\ldots>\mu_{\textrm{definitely negative}}$}. For the SUN dataset, we defined two types of attribute relationships, \emph{i.e.} {$l_{ij}\in\{\textrm{negative},\textrm{positive}\}$}. We manually annotated 18 probably positive relationships, 549 not-related relationships, 21 probably negative relationships, and 9 definitely negatively relationships in the CelebA dataset. In the SUN dataset, we labeled a total of 83 positive relationships and 63 negative relationships.

\textbf{Experiment 1, mining potentially biased attribute representations: } We trained two CNNs using images from the CelebA dataset and those from the SUN dataset, respectively. Then, we diagnosed attribute representations of the CNNs. Fig.~\ref{fig:graph} compares the mined and the ground-truth attribute relationships.

Our method is not sensitive to a small number of errors in ground-truth relationship annotations. It is because as shown in Fig.~\ref{fig:graph}, for each attribute, we compute multiple pairwise relationships between this attribute and other attributes. A single inaccurate relationship will not significantly affect the result. Similarly, because we calculate the KL divergence among all training images, KL divergence results in Fig.~\ref{fig:graph_kl} are robust to noise and appearance variations in specific images. The low error rate of attribute estimation is not necessarily equivalent to good representations. Error rates for the ``wearing lipstick'' and ``double chin'' attributes are only 8.1\% and 4.5\%, which are lower than the average error 11.1\%. However, these two attributes have the top-4 representation biases.

\begin{figure*}[t]
\centering
\includegraphics[width=0.7\linewidth]{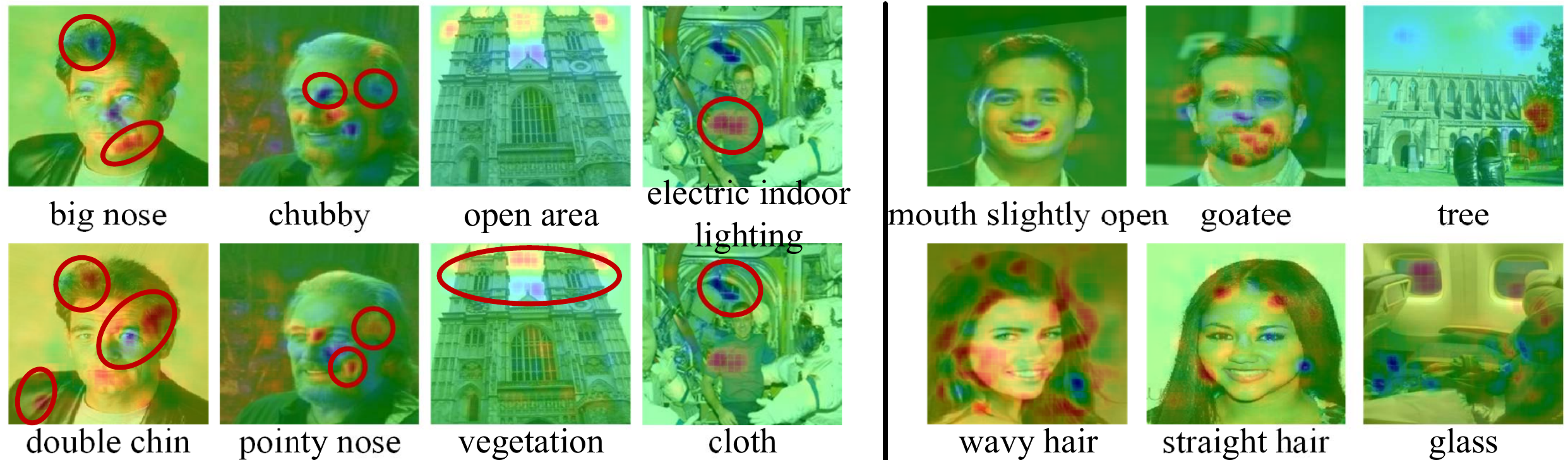}
\caption{Good and bad representations of CNNs. (left) For a failure mode, we show heat maps of inference patterns of the two attributes in the failure mode. Red/blue colors on faces show the patterns that increase/decrease the attribute score. Red circles indicate incorrect representations, where the CNN mistakenly uses incorrect inference patterns to represent the attribute. (right) Well learned attribute representations.}
\label{fig:example}
\end{figure*}

As shown in Fig.~\ref{fig:example}, when the CNN uses patterns in incorrect positions to represent the attribute, we will probably obtain a significant KL divergence.

\textbf{Experiment 2, testing the proposed method on manually biased datasets: } In this experiment, we manually biased training sets to learn CNNs. We used our method to explore the relationship between the dataset bias and the representation bias in the CNNs.

From each of the CelebA and the SUN datasets, we randomly selected 10 pairs of attributes. Then, for each pair of attributes, {$(A_{i},A_{j})$}, we biased the distribution of {$A_{i}$} and {$A_{j}$}'s ground-truth annotations to produce a new training set, as follows. Given a parameter $\tau$ ({$0\leq\tau\leq1$}) that denotes the bias level, we randomly removed {$\tau\cdot N_{Y_{i}^{*}Y_{j}^{*}\!<\!0}$} samples from all samples whose ground-truth annotations {$Y_{i}^{*}$} and {$Y_{j}^{*}$} were opposite, where {$N_{Y_{i}^{*}Y_{j}^{*}\!<\!0}$} denotes the number of samples that satisfied {$Y_{i}^{*}Y_{j}^{*}\!<\!0$}.

Initially, for each pair of attributes {$(A_{i},A_{j})$}, we generated a fully biased training set with {$\tau=1$}, and our method mined a significant KL divergence of {${\bf KL}_{A_{i}A_{j}}$}. We then gradually added samples with {$Y_{i}^{*}Y_{j}^{*}\!<\!0$} to reduce the dataset bias $\tau$, and learned new CNNs based on the new training sets. Fig.~\ref{fig:kl} shows the decrease of the KL divergence when the dataset bias $\tau$ was reduced. Given each of the 10 pairs of attributes, we generated four biased datasets by applying four values of {$\tau\in\{0.25,0.5,0.75,1.0\}$}. In this way, we obtained 40 biased CelebA datasets ($\tau\in\{0.25,0.5,0.75,1.0\}$) and another 30 biased SUN Attribute datasets ($\tau\in\{0.5,0.75,1.0\}$) to learn 70 CNNs. Fig.~\ref{fig:kl} shows KL divergences mined from these CNNs.

The experiment demonstrates that large KL divergences successfully reflected potentially biased representations, but the level of annotation bias was \textbf{not} proportional to the level of representation bias. When we reduced the annotation bias $\tau$, the corresponding KL divergence usually decreased. At the meanwhile, CNN representations had different sensitiveness to different types of annotation bias. Small bias \emph{w.r.t.} some pairs of attributes (\emph{e.g.} \textit{heavy makeup} and \textit{pointy nose}) led to huge representation bias. Whereas, the CNN was robust to annotation biases of other pairs of attributes. For example, it was easy for the CNN to extract correct inference patterns for \textit{male} and \textit{oval face}, so small annotation bias of these two attributes did not cause a significant representation bias (see the lowest gray curve in Fig.~\ref{fig:kl}(left)).

\begin{figure*}[t]
\centering
\includegraphics[width=\linewidth]{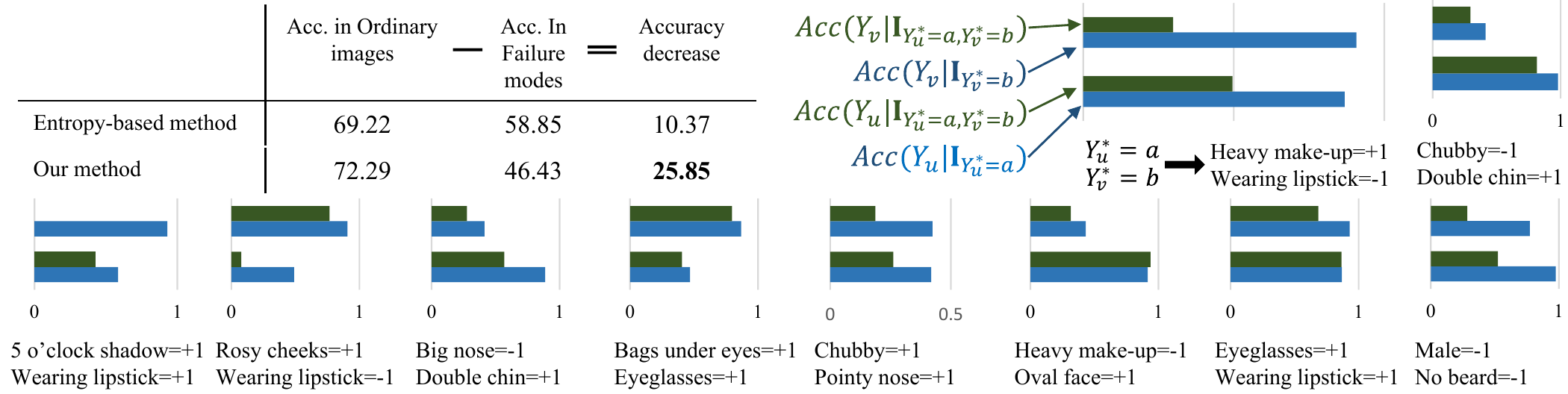}
\caption{Top-10 failure modes of a CNN trained using the CelebA dataset. The top-left table shows average accuracy decrease caused by failure modes. Each sub-figure illustrates the accuracy decrease caused by a failure mode {$(Y_{u}^{*}=a,Y_{v}^{*}=b)$} mined by our method.}
\label{fig:mode}
\end{figure*}

\textbf{Experiment 3, the discovery of blind spots and failure modes: } In this experiment, we mined blind spots and failure modes. We obtained five blind spots from the CNN for the CelebA dataset, \emph{i.e.} the CNN did not encode positive relationships between \textit{attractive} and each of \textit{earrings}, \textit{necktie}, and \textit{necklace}, the negative relationship between \textit{chubby} and \textit{oval face}, and the negative relationship between \textit{bangs} and \textit{wearing hat}. We list blind spots with top-10 KL divergences that were mined from the CNN for the SUN Attribute database in Table~\ref{tab:blind}. For example, the CNN did not learn a strong negative relationship between \textit{man-made} and \textit{vegetation} as expected, because \textit{man-made} and \textit{vegetation} co-exist in many training images. This is a typical dataset bias, and the dataset should contain more images that have only one of the two attributes.

We mined failure modes with top-$N$ KL divergences. We compared our method with an entropy-based method for the discovery of failure modes. The competing method only used distributions of ground-truth annotations to predict potential failure modes of the CNN. For each pair of attributes {$(A_{i},A_{j})$}, its failure mode was defined as the mode corresponding to the least training samples among all the four mode candidates {$(Y_{i}^{*}\!=\!+1,Y_{j}^{*}\!=\!+1)$}, {$(Y_{i}^{*}\!=\!+1,Y_{j}^{*}\!=\!-1)$}, {$(Y_{i}^{*}\!=\!-1,Y_{j}^{*}\!=\!+1)$}, {$(Y_{i}^{*}\!=\!-1,Y_{j}^{*}\!=\!-1)$}. The significance of the failure mode was computed as the entropy of the joint distribution of {$(Y_{i}^{*},Y_{j}^{*})$} among training samples{\textcolor{red}{\footnotemark[4]}}. Then, we selected failure modes with top-$N$ entropies as results. To evaluate the effectiveness of failure modes, we tested the CNN using testing images. Let {$(Y_{u}^{*}\!=\!a,Y_{v}^{*}\!=\!b)$} ({$a,b\in\{-1,+1\}$}) be a failure mode and {$Acc(Y_{u}|{\bf I}_{Y_{u}^{*}\!=\!a})$} denote the accuracy for estimating {$A_{u}$} on testing images with {$Y_{u}^{*}\!=\!a$}. Then, {$[Acc(Y_{u}|{\bf I}_{Y_{u}^{*}=a})+Acc(Y_{v}|{\bf I}_{Y_{v}^{*}=b})]/2$} measures the accuracy on ordinary images, and {$[Acc(Y_{u}|{\bf I}_{Y_{u}^{*}=a,Y_{v}^{*}=b})+Acc(Y_{v}|{\bf I}_{Y_{u}^{*}=a,Y_{v}^{*}=b})]/2$} measures the accuracy on images with the failure modes. Fig.~\ref{fig:mode} compares the accuracy decrease caused by the top-10 failure modes mined by our method and the accuracy decrease caused by the top-10 failure modes produced by the competing method. Table~\ref{tab:failureMode} shows accuracy decreases caused by different numbers of failure modes. It showed that our method extracted more reliable failure modes. Fig.~\ref{fig:example} further visualizes biased representations corresponding to some failure modes. Intuitively, failure modes in Fig.~\ref{fig:mode} also fit biased co-appearance of two attributes among training images.

\textbf{Justification of the methodology:}{\verb| |} Incorrect representations are usually caused by dataset bias and the over-fitting problem. For example, if {$A_1$} often has a positive annotation when {$A_2$} is labeled positive (or negative), the CNN may use {$A_2$}'s features as a contextual information to describe {$A_1$}. However, in real applications, it is difficult to predict whether the algorithm will suffer from such dataset bias before the learning process. For example, when the conditional distribution {$P(Y_1^{*}|Y_2^{*}>0)$} is biased (\emph{e.g.} {$P(Y_1^{*}>0|Y_2^{*}>0)>P(Y_1^{*}<0|Y_2^{*}>0)$}) but {$P(Y_1^{*}|Y_2^{*}<0)$} has a balance distribution, it is difficult to predict whether the CNN will consider {$A_1$} and {$A_2$} are positively related to each other.

Let us discuss two toy cases of this problem for simplification. Let us assume that the CNN mainly extracts common features from positive samples with {$Y_2^{*}>0$} to represent {$A_2$}, and regards negative samples with {$Y_2^{*}<0$} as random samples without sharing common features. In this case, the conditional distribution {$P(Y_1^{*}|Y_2^{*}>0)$} will probably control the relationships between {$A_1$} and {$A_2$}. Whereas, if the CNN mainly extracts features from negative samples with {$Y_2^{*}<0$} to represent {$A_2$}, then the attribute relationship will not be sensitive to the conditional distribution {$P(Y_1^{*}|Y_2^{*}>0)$}.

Therefore, as shown in Fig.~\ref{fig:mode} and Table~\ref{tab:failureMode}, our method is more effective in the discovery of failure modes than the method based on the entropy of annotation distributions.

\begin{table}[t]
\resizebox{1.0\linewidth}{!}{\begin{tabular}{l|l|ccc}
\hline
\multicolumn{5}{c}{CelebA dataset}\\
\hline
& &\!\!\! {\footnotesize Accuracy in} \!\!\!&\!\!\! {\footnotesize Accuracy in} \!\!\!&\!\!\! {\footnotesize Decrease of}\!\!\!\\
& &\!\!\! {\footnotesize ordinary images} \!\!\!&\!\!\! {\footnotesize failure modes} \!\!\!&\!\!\! {\footnotesize accuracy}\!\!\!\\
\hline
\!\!\!top-5 & {\footnotesize Entropy-based} & 74.10 & 60.37 & \color{blue}{13.73}\\
& {\footnotesize Our method} & 73.81 & 40.22 & \color{blue}{\bf33.59}\\
\hline
\!\!\!top-10 & {\footnotesize Entropy-based} & 69.22 & 58.85 & \color{blue}{10.37}\\
& {\footnotesize Our method} & 72.29 & 46.43 & \color{blue}{\bf25.85}\\
\hline
\!\!\!top-15 & {\footnotesize Entropy-based} & 67.49 & 56.44 & \color{blue}{11.05}\\
& {\footnotesize Our method} & 68.05 & 47.95 & \color{blue}{\bf20.10}\\
\hline
\!\!\!top-20 & {\footnotesize Entropy-based} & 68.06 & 57.32 & \color{blue}{10.73}\\
& {\footnotesize Our method} & 66.94 & 46.57 & \color{blue}{\bf20.37}\\
\hline
\!\!\!top-25 & {\footnotesize Entropy-based} & 68.24 & 59.79 & \color{blue}{8.45}\\
& {\footnotesize Our method} & 67.06 & 49.23 & \color{blue}{\bf17.83}\\
\hline
\multicolumn{5}{c}{SUN Attribute database}\\
\hline
\!\!\!top-40 & {\footnotesize Entropy-based} & 63.36 & 35.89 & \color{blue}{27.47}\\
& {\footnotesize Our method} & 68.65 & 38.98 & \color{blue}{\bf29.68}\\
\hline
\!\!\!top-50 & {\footnotesize Entropy-based} & 59.29 & 35.62 & \color{blue}{23.67}\\
& {\footnotesize Our method} & 65.73 & 38.86 & \color{blue}{\bf26.87}\\
\hline
\end{tabular}}
\caption{Average accuracy decrease caused by top-$N$ failure modes, which were mined from the CNN for the CelebA dataset ($N=5,10,15,20,25$) and the CNN for the SUN Attribute database ($N=40,50$). We compare the entropy-based method with our method.}
\label{tab:failureMode}
\end{table}

\begin{table*}
\centering
\begin{tabular}{ccl}
\hline
$\#$ & ${\bf KL}_{A_{i}A_{j}}$ & Description of blind spots\\
\hline
1 & 1.050 & negative relationship between \textit{foliage} and \textit{man-made}\\
2 & 1.048 & negative relationship between \textit{leaves} and \textit{man-made}\\
3 & 1.020 & negative relationship between \textit{trees} and \textit{mostly vertical components}\\
4 & 0.982 & positive relationship between \textit{cloth} and \textit{matte}\\
5 & 0.787 & negative relationship between \textit{dry} and \textit{enclosed area}\\
6 & 0.776 & positive relationship between \textit{dry} and \textit{open area}\\
7 & 0.774 & negative relationship between \textit{foliage} and \textit{mostly vertical components}\\
8 & 0.770 & negative relationship between \textit{leaves} and \textit{mostly vertical components}\\
9 & 0.767 & positive relationship between \textit{metal} and \textit{man-made}\\
10 & 0.751 & negative relationship between \textit{dry} and \textit{man-made}\\
\hline
\end{tabular}
\caption{Blind spots mined from the CNN that is trained using the SUN Attribute database. We list blind spots with the top-10 KL divergences in the table.}
\label{tab:blind}
\end{table*}

\section{Summary and discussion}

In this paper, we have designed a method to explore inner conflicts inside representations of a pre-trained CNN without given any additional testing samples. This study focuses on an essential yet commonly ignored issue in artificial intelligence, \emph{i.e.} how can we ensure the CNN learns what we expect it to learn. When there is a dataset bias, the CNN may use unreliable contexts to represent an attribute. Our method mines failure modes of a CNN, which can potentially guide the collection of new training samples. Experiments have demonstrated the high correlations between the mined KL divergences and dataset bias and shown the effectiveness in the discovery of failure modes.


In this paper, we used Gaussian distributions to approximate ground-truth distributions of attribute relationships to simplify the story. However, our method can be extended and use more complex distributions according to each specific application. In addition, it is difficult to say all discovered representation biases are ``definitely'' incorrect representations. For example, the CNN may use \textit{rosy cheeks} to identify the \textit{wearing lipstick} attribute, but these two attributes are ``indirectly'' related to each other. It is problematic to annotate the two attributes are either positively related or not related to each other. The \textit{wearing necktie} attribute is directly related to the \textit{male} attribute, but is indirectly related to the \textit{mustache} attribute, because the necktie and the mustache describe different parts of the face. If we label \textit{wearing necktie} is not related to \textit{mustache}, then our method will examine whether the CNN uses mustache as contexts to describe the necktie. Similarly, if we consider such an indirect relationship as reliable contexts, we can simply annotate a positive relationship between \textit{necktie} and \textit{mustache}. Moreover, if neither the ``not-related'' relationship nor the positive relationship between the two attributes is trustworthy, we can simply ignore such relationships to avoid the risk of incorrect ground truth. In the future work, we would encode ground-truth attribute relationships as a prior into the end-to-end learning of CNNs, in order to achieve more reasonable representations.


\section*{Acknowledgement}
This work is supported by ONR MURI project N00014-16-1-2007 and DARPA XAI Award N66001-17-2-4029, and NSF IIS 1423305.

\bibliographystyle{aaai}
\bibliography{TheBib}

\begin{thebibliography}{}

\bibitem[\protect\citeauthoryear{Adler \bgroup et al\mbox.\egroup
  }{2016}]{blackBoxKeyFeature}
Adler, P.; Falk, C.; Friedler, S.~A.; Rybeck, G.; Scheidegger, C.; Smith, B.;
  and Venkatasubramanian, S.
\newblock 2016.
\newblock Auditing black-box models for indirect influence.
\newblock {\em In ICDM}.

\bibitem[\protect\citeauthoryear{Aubry and Russell}{2015}]{CNNVisualization_5}
Aubry, M., and Russell, B.~C.
\newblock 2015.
\newblock Understanding deep features with computer-generated imagery.
\newblock {\em In ICCV}.

\bibitem[\protect\citeauthoryear{Bansal, Farhadi, and
  Parikh}{2014}]{failureMode}
Bansal, A.; Farhadi, A.; and Parikh, D.
\newblock 2014.
\newblock Towards transparent systems: Semantic characterization of failure
  modes.
\newblock {\em In ECCV}.

\bibitem[\protect\citeauthoryear{Deng \bgroup et al\mbox.\egroup
  }{2014}]{labelGraph}
Deng, J.; Ding, N.; Jia, Y.; Frome, A.; Murphy, K.; Bengio, S.; Li, Y.; Neven,
  H.; and Adam, H.
\newblock 2014.
\newblock Large-scale object classification using label relation graphs.
\newblock {\em In ECCV}.

\bibitem[\protect\citeauthoryear{Dosovitskiy and Brox}{2016}]{FeaVisual}
Dosovitskiy, A., and Brox, T.
\newblock 2016.
\newblock Inverting visual representations with convolutional networks.
\newblock {\em In CVPR}.

\bibitem[\protect\citeauthoryear{Farhadi \bgroup et al\mbox.\egroup
  }{2009}]{attribute}
Farhadi, A.; Endres, I.; Hoiem, D.; and Forsyth, D.
\newblock 2009.
\newblock Describing objects by their attributes.
\newblock {\em In CVPR}.

\bibitem[\protect\citeauthoryear{Fong and Vedaldi}{2017}]{visualCNN_grad}
Fong, R.~C., and Vedaldi, A.
\newblock 2017.
\newblock Interpretable explanations of black boxes by meaningful perturbation.
\newblock {\em In arXiv:1704.03296v1}.

\bibitem[\protect\citeauthoryear{Goyal \bgroup et al\mbox.\egroup
  }{2016}]{interpretVQA_grad}
Goyal, Y.; Mohapatra, A.; Parikh, D.; and Batra, D.
\newblock 2016.
\newblock Towards transparent ai systems: Interpreting visual question
  answering models.
\newblock {\em In arXiv:1608.08974v2}.

\bibitem[\protect\citeauthoryear{Hu \bgroup et al\mbox.\egroup
  }{2016}]{LogicRuleNetwork}
Hu, Z.; Ma, X.; Liu, Z.; Hovy, E.; and Xing, E.~P.
\newblock 2016.
\newblock Harnessing deep neural networks with logic rules.
\newblock {\em In arXiv:1603.06318v2}.

\bibitem[\protect\citeauthoryear{Koh and Liang}{2017}]{neuralGap}
Koh, P., and Liang, P.
\newblock 2017.
\newblock Understanding black-box predictions via influence functions.
\newblock {\em In arXiv preprint, arXiv:1703.04730}.

\bibitem[\protect\citeauthoryear{Krizhevsky, Sutskever, and
  Hinton}{2012}]{CNNImageNet}
Krizhevsky, A.; Sutskever, I.; and Hinton, G.
\newblock 2012.
\newblock Imagenet classification with deep convolutional neural networks.
\newblock {\em In NIPS}.

\bibitem[\protect\citeauthoryear{Lakkaraju \bgroup et al\mbox.\egroup
  }{2017}]{banditUnknown}
Lakkaraju, H.; Kamar, E.; Caruana, R.; and Horvitz, E.
\newblock 2017.
\newblock Identifying unknown unknowns in the open world: Representations and
  policies for guided exploration.
\newblock {\em In AAAI}.

\bibitem[\protect\citeauthoryear{Liu \bgroup et al\mbox.\egroup
  }{2015}]{CelebA}
Liu, Z.; Luo, P.; Wang, X.; and Tang, X.
\newblock 2015.
\newblock Deep learning face attributes in the wild.
\newblock {\em In ICCV}.

\bibitem[\protect\citeauthoryear{Liu, Shen, and van~den
  Hengel}{2015}]{CNNFeatureMining}
Liu, L.; Shen, C.; and van~den Hengel, A.
\newblock 2015.
\newblock The treasure beneath convolutional layers: Cross-convolutional-layer
  pooling for image classification.
\newblock {\em In CVPR}.

\bibitem[\protect\citeauthoryear{Long and Hua}{2015}]{Active2}
Long, C., and Hua, G.
\newblock 2015.
\newblock Multi-class multi-annotator active learning with robust gaussian
  process for visual recognition.
\newblock {\em In ICCV}.

\bibitem[\protect\citeauthoryear{Mahendran and
  Vedaldi}{2015}]{CNNVisualization_2}
Mahendran, A., and Vedaldi, A.
\newblock 2015.
\newblock Understanding deep image representations by inverting them.
\newblock {\em In CVPR}.

\bibitem[\protect\citeauthoryear{Patterson \bgroup et al\mbox.\egroup
  }{2014}]{SUNAttr}
Patterson, G.; Xu, C.; Su, H.; and Hays, J.
\newblock 2014.
\newblock The sun attribute database: Beyond categories for deeper scene
  understanding.
\newblock {\em In IJCV}.

\bibitem[\protect\citeauthoryear{Ribeiro, Singh, and Guestrin}{2016}]{trust}
Ribeiro, M.~T.; Singh, S.; and Guestrin, C.
\newblock 2016.
\newblock ``why should i trust you?'' explaining the predictions of any
  classifier.
\newblock {\em In KDD}.

\bibitem[\protect\citeauthoryear{Ross, Hughes, and
  Doshi-Velez}{2017}]{rightReason}
Ross, A.~S.; Hughes, M.~C.; and Doshi-Velez, F.
\newblock 2017.
\newblock Right for the right reasons: Training differentiable models by
  constraining their explanations.
\newblock {\em In arXiv:1703.03717v1}.

\bibitem[\protect\citeauthoryear{Selvaraju \bgroup et al\mbox.\egroup
  }{2017}]{visualCNN_grad_2}
Selvaraju, R.~R.; Cogswell, M.; Das, A.; Vedantam, R.; Parikh, D.; and Batra,
  D.
\newblock 2017.
\newblock Grad-cam: Visual explanations from deep networks via gradient-based
  localization.
\newblock {\em In arXiv:1610.02391v3}.

\bibitem[\protect\citeauthoryear{Simon and Rodner}{2015}]{ObjectDiscoveryCNN_2}
Simon, M., and Rodner, E.
\newblock 2015.
\newblock Neural activation constellations: Unsupervised part model discovery
  with convolutional networks.
\newblock {\em In ICCV}.

\bibitem[\protect\citeauthoryear{Simon, Rodner, and
  Denzler}{2014}]{CNNSemanticPart}
Simon, M.; Rodner, E.; and Denzler, J.
\newblock 2014.
\newblock Part detector discovery in deep convolutional neural networks.
\newblock {\em In ACCV}.

\bibitem[\protect\citeauthoryear{Simonyan, Vedaldi, and
  Zisserman}{2014}]{CNNVisualization_3}
Simonyan, K.; Vedaldi, A.; and Zisserman, A.
\newblock 2014.
\newblock Deep inside convolutional networks: Visualising image classification
  models and saliency maps.
\newblock {\em In ICLR Workshop}.

\bibitem[\protect\citeauthoryear{Suh, Zhu, and
  Amershi}{2016}]{MachineTeachingActive}
Suh, J.; Zhu, X.; and Amershi, S.
\newblock 2016.
\newblock The label complexity of mixed-initiative classifier training.
\newblock {\em In ICML}.

\bibitem[\protect\citeauthoryear{Torralba and Efros}{2011}]{datasetBias}
Torralba, A., and Efros, A.
\newblock 2011.
\newblock Unbiased look at dataset bias.
\newblock {\em In CVPR}.

\bibitem[\protect\citeauthoryear{Vijayanarasimhan and
  Grauman}{2011}]{activeDPMGrauman}
Vijayanarasimhan, S., and Grauman, K.
\newblock 2011.
\newblock Large-scale live active learning: Training object detectors with
  crawled data and crowds.
\newblock {\em In CVPR}.

\bibitem[\protect\citeauthoryear{Wu and Zhu}{2011}]{wu2011numerical}
Wu, T., and Zhu, S.-C.
\newblock 2011.
\newblock A numerical study of the bottom-up and top-down inference processes
  in and-or graphs.
\newblock {\em International journal of computer vision} 93(2):226--252.

\bibitem[\protect\citeauthoryear{Wu, Xia, and Zhu}{2007}]{wu2007compositional}
Wu, T.-F.; Xia, G.-S.; and Zhu, S.-C.
\newblock 2007.
\newblock Compositional boosting for computing hierarchical image structures.
\newblock {\em In CVPR}.

\bibitem[\protect\citeauthoryear{Yang, Wu, and Zhu}{2009}]{yang2009evaluating}
Yang, X.; Wu, T.; and Zhu, S.-C.
\newblock 2009.
\newblock Evaluating information contributions of bottom-up and top-down
  processes.
\newblock {\em ICCV}.

\bibitem[\protect\citeauthoryear{Zeiler and Fergus}{2014}]{CNNVisualization_1}
Zeiler, M.~D., and Fergus, R.
\newblock 2014.
\newblock Visualizing and understanding convolutional networks.
\newblock {\em In ECCV}.

\bibitem[\protect\citeauthoryear{Zhang \bgroup et al\mbox.\egroup
  }{2014}]{failureMode2}
Zhang, P.; Wang, J.; Farhadi, A.; Hebert, M.; and Parikh, D.
\newblock 2014.
\newblock Predicting failures of vision systems.
\newblock {\em In CVPR}.

\bibitem[\protect\citeauthoryear{Zhang \bgroup et al\mbox.\egroup
  }{2016}]{CNNAoG}
Zhang, Q.; Cao, R.; Wu, Y.~N.; and Zhu, S.-C.
\newblock 2016.
\newblock Growing interpretable part graphs on convnets via multi-shot
  learning.
\newblock {\em In AAAI}.

\bibitem[\protect\citeauthoryear{Zhang \bgroup et al\mbox.\egroup
  }{2017a}]{interactiveAOG_arXiv}
Zhang, Q.; Cao, R.; Zhang, S.; Edmonds, M.; Wu, Y.; and Zhu, S.-C.
\newblock 2017a.
\newblock Interactively transferring cnn patterns for part localization.
\newblock {\em In arXiv:1708.01783}.

\bibitem[\protect\citeauthoryear{Zhang \bgroup et al\mbox.\egroup
  }{2017b}]{DeepQA}
Zhang, Q.; Cao, R.; Wu, Y.~N.; and Zhu, S.-C.
\newblock 2017b.
\newblock Mining object parts from cnns via active question-answering.
\newblock {\em In CVPR}.

\bibitem[\protect\citeauthoryear{Zhang \bgroup et al\mbox.\egroup
  }{2018}]{explanatoryGraph}
Zhang, Q.; Cao, R.; Shi, F.; Wu, Y.; and Zhu, S.-C.
\newblock 2018.
\newblock Interpreting cnn knowledge via an explanatory graph.
\newblock {\em In AAAI}.

\bibitem[\protect\citeauthoryear{Zhou \bgroup et al\mbox.\egroup
  }{2015}]{CNNSemanticDeep}
Zhou, B.; Khosla, A.; Lapedriza, A.; Oliva, A.; and Torralba, A.
\newblock 2015.
\newblock Object detectors emerge in deep scene cnns.
\newblock {\em In ICRL}.

\bibitem[\protect\citeauthoryear{Zhou \bgroup et al\mbox.\egroup
  }{2016}]{CNNSemanticDeep2}
Zhou, B.; Khosla, A.; Lapedriza, A.; Oliva, A.; and Torralba, A.
\newblock 2016.
\newblock Learning deep features for discriminative localization.
\newblock {\em In CVPR}.

\end{thebibliography}

\end{document}